# Clinical Evaluation of a Tongue-Controlled Wrist Abduction-Adduction Assistance in a 6-DoF Upper-Limb Exoskeleton for Individuals with ALS and SCI

Juwairiya S. Khan[1,6], Mostafa Mohammadi[1], Alexander L. Ammitzbøll[2], Ellen-Merete Hagen[2,4], Jakob Blicher[3] Izabella Obál[3], Ana S. S. Cardoso[1], Oguzhan Kirtas[5], Rasmus L. Kæseler[1], John Rasmussen[6], and Lotte N.S. Andreasen Struijk[1]

***Abstract*—Upper-limb exoskeletons (ULEs) have the potential to restore functional independence in individuals with severe motor impairments; however, the clinical relevance of wrist degrees of freedom (DoF), particularly abduction-adduction (Ab-Ad), remains insufficiently evaluated. This study investigates the functional and user-perceived impact of wrist Ab-Ad assistance during two activities of daily living (ADLs). Wrist Ab-Ad assistance in a tongue-controlled 6-DoF ULE, EXOTIC[2], was evaluated in a within-subject study involving one individual with amyotrophic lateral sclerosis and five individuals with spinal cord injury. Participants performed drinking and scratch stick leveling tasks with EXOTIC[2] under two conditions: with and without wrist Ab-Ad assistance. Outcome measure included task success, task completion time, kinematic measures, and a usability questionnaire capturing comfort, functional perception, and acceptance. Enabling wrist Ab-Ad improved task success rates across both ADLs, with consistent reductions in spillage (from 77.8% spillages to 22.2%) and failed placements (from 66.7% to 16.7%). Participants utilized task-specific subsets of the available wrist range of motion, indicating that effective control within functional ranges was more critical than maximal joint excursion. Questionnaire responses indicated no increase in discomfort with the additional DoF and reflected perceived improvements in task performance. In conclusion, wrist Ab-Ad assistance enhances functional task performance in assistive exoskeleton use without compromising user comfort. However, its effectiveness depends on task context, control usability, and individual user strategies. This study provides clinically relevant, user-centered evidence supporting the inclusion of wrist Ab-Ad in ULEs, emphasizing the importance of balancing functional capability with usability in assistive device design.**

**This work has been submitted to the IEEE Transactions on Neural Systems and Rehabilitation Engineering for possible publication. Copyright may be transferred without notice, after which this version may no longer be accessible.**

*Research supported by Aage og Johanne Louis-Hansens Fond.

[1]Juwairiya Siraj Khan (corresponding author), Mostafa Mohammadi, Ana S. S. Cardoso, Rasmus L. Kæseler and Lotte N.S. Andreasen. Struijk are with the Neurorehabilitation Robotics and Engineering, Center for Rehabilitation Robotics, Department of Health Science and Technology, Aalborg University, Gistrup, Denmark. (jsikh@hst.aau.dk , mostafa@hst.aau.dk, assc@hst.aau.dk , rlk@hst.aau.dk, naja@hst.aau.dk ).

[2]ALA (alexnl@rm.dk) and EMH (elhage@rm.dk) are with the Spinal Cord Injury Centre of Western Denmark, Regional Hospital of Viborg Denmark

[3] JB (j.blicher@rn.dk) and IO (i.obal@rn.dk) are with Aalborg University Hospital, Department of Neurology, 9000 Aalborg, Denmark

[4]EMH (elhage@rm.dk) is also with the Department of Clinical Medicine, Aarhus University, Denmark and Department of Brain Repair and Rehabilitation, UCL Queen Square Institute of Neurology, University College London, UK

5OK (okr@create.aau.dk) is with the Department of Architecture, Design and Media Technology, Aalborg University, Aalborg 9000, Denmark.

[6]Juwairiya Siraj Khan and John Rasmussen are with the Department of Materials and Production, Aalborg University, Aalborg East, Denmark (e-mail: jsikh@mp.aau.dk , jr@mp.aau.dk)

## I. INTRODUCTION

NEUROLOGICAL conditions leading to severe upper-limb impairment affect millions of individuals worldwide. Spinal cord injury (SCI) alone affects more than 15 million people globally, with a significant proportion resulting in tetraplegia and loss of arm and hand function [1]. Similarly, amyotrophic lateral sclerosis (ALS), with an incidence of approximately 1-2 cases per 100,000 individuals per year, leads to progressive degeneration of motor neurons and eventual paralysis [2], [3].

The loss of upper-limb function is one of the most debilitating impairments in both ALS and SCI populations, directly limiting independence in activities of daily living (ADLs) such as eating, drinking, personal hygiene, and object manipulation. As a result, individuals with tetraplegia often become highly dependent on caregivers for routine tasks, reducing autonomy and quality of life while increasing long-term care burden. This highlights the need for clinically relevant assistive solutions that restore functional independence in ADLs rather than enabling only isolated joint movements. Assistive technologies, particularly wearable robotic exoskeletons, have been proposed to address these challenges [4].

Upper-limb exoskeletons have advanced considerably in recent years, with systems demonstrating assistance in reaching and manipulation tasks [5]. Several designs incorporate multi-degree-of-freedom (DoF) actuation of the shoulder, elbow, and forearm, along with control interfaces such as electromyography (EMG), brain-computer interfaces (BCIs), or alternative human-machine interfaces [6], [7]. However, most existing systems either target users with residual motor function or lack sufficient number of assisted DOFs. As a result, their effectiveness in supporting functional ADLs in individuals with severe or complete upper-limb paralysis remains insufficiently validated in clinically relevant contexts [8][6].

A key limitation in current upper-limb exoskeleton designs is the lack of comprehensive wrist functionality, particularly wrist abduction-adduction (Ab-Ad) or radial-ulnar deviation [9], [10], [11]. While flexion-extension and forearm rotation are commonly implemented, radial-ulnar deviation is often omitted due to challenges related to mechanical complexity, alignment, and added weight [9], [12]. Nevertheless, biomechanical studies highlight the importance of wrist motion in enabling efficient object manipulation and compensatory strategies during ADLs [13], [14], [15]. Despite this, many designs prioritize proximal joints like shoulder and elbow and limited degrees of freedom over a full arm with both proximal and distal joints, with relatively few systems incorporating wrist-level assistance or providing clinical evidence of functional benefits of wrist in ADLs for individuals with severe motor impairments[6], [16].

In addition to design limitations, there is a broader gap in user-centered clinical evaluation of wrist joints such as Ab-Ad in target populations such as ALS and high-level SCI [6], [17], [18]. Many studies rely on healthy participants or individuals with mild impairments, which limits the generalizability of findings [6]. Furthermore, evaluations often focus on kinematic performance or isolated movements rather than task-level ADL outcomes combined with user-reported measures such as comfort, usability, and perceived usefulness [6], [8]. Consequently, it remains unclear whether increasing system complexity, such as adding additional degrees of freedom, translates into clinically meaningful improvements in real-world assistive use, as higher mechanical and control complexity does not necessarily result in improved functional outcomes or usability in assistive robotic systems [19], [20].

To address these gaps, this exploratory study presents a clinical, user-centered evaluation of wrist Ab-Ad assistance in the EXOTIC$^{2}$ [15] exoskeleton, which is an extension of the 5-Dof upper limb exoskeleton, EXOTIC [18] exoskeleton, by integrating a wrist Ab-Ad joint to enable wrist joint assistance. The system is designed for individuals with tetraplegia [21] and is controlled by tongue motions through a non-invasive tongue computer interface (nTCI) [22], [23]. The evaluation of the wrist joint was conducted with individuals with tetraplegia caused by ALS and SCI performing ADLs with and without wrist abduction-adduction assistance.

Thus, the main contribution of this work is a comparative clinical evaluation of ADL performance with and without wrist abduction-adduction support in individuals with severe motor impairments. along with subjective user-centered measures of usability, comfort, and perceived benefit. By focusing on real end users and task-level outcomes, this work aims to bridge the gap between engineering design and clinically meaningful assistive function.

## II. Materials and Methods

### A. System Description

#### 1) EXOTIC$^{2}$ Exoskeleton

The study was conducted using a wearable upper-limb exoskeleton (EXOTIC$^{2}$) configured with six actuated degrees of freedom (DoFs) [15] (Fig. 1), to support functional arm movements in individuals with severe motor impairments. The system provides actuation for shoulder flexion/extension and internal/external rotation, elbow flexion/extension, forearm pronation/supination, wrist abduction/adduction, and hand opening/closing [ [15], [18]. Shoulder, elbow and forearm joints are actuated through rigid transmission mechanisms to ensure stability and torque generation, while wrist Ab-Ad is implemented using tendon-driven actuation to reduce inertia at the hand [9]. Grasp assistance is provided through a cable-driven soft robotic glove (CarbonHand, Bioservo, Sweden), enabling hand closure, while passive elastic elements assist hand opening.

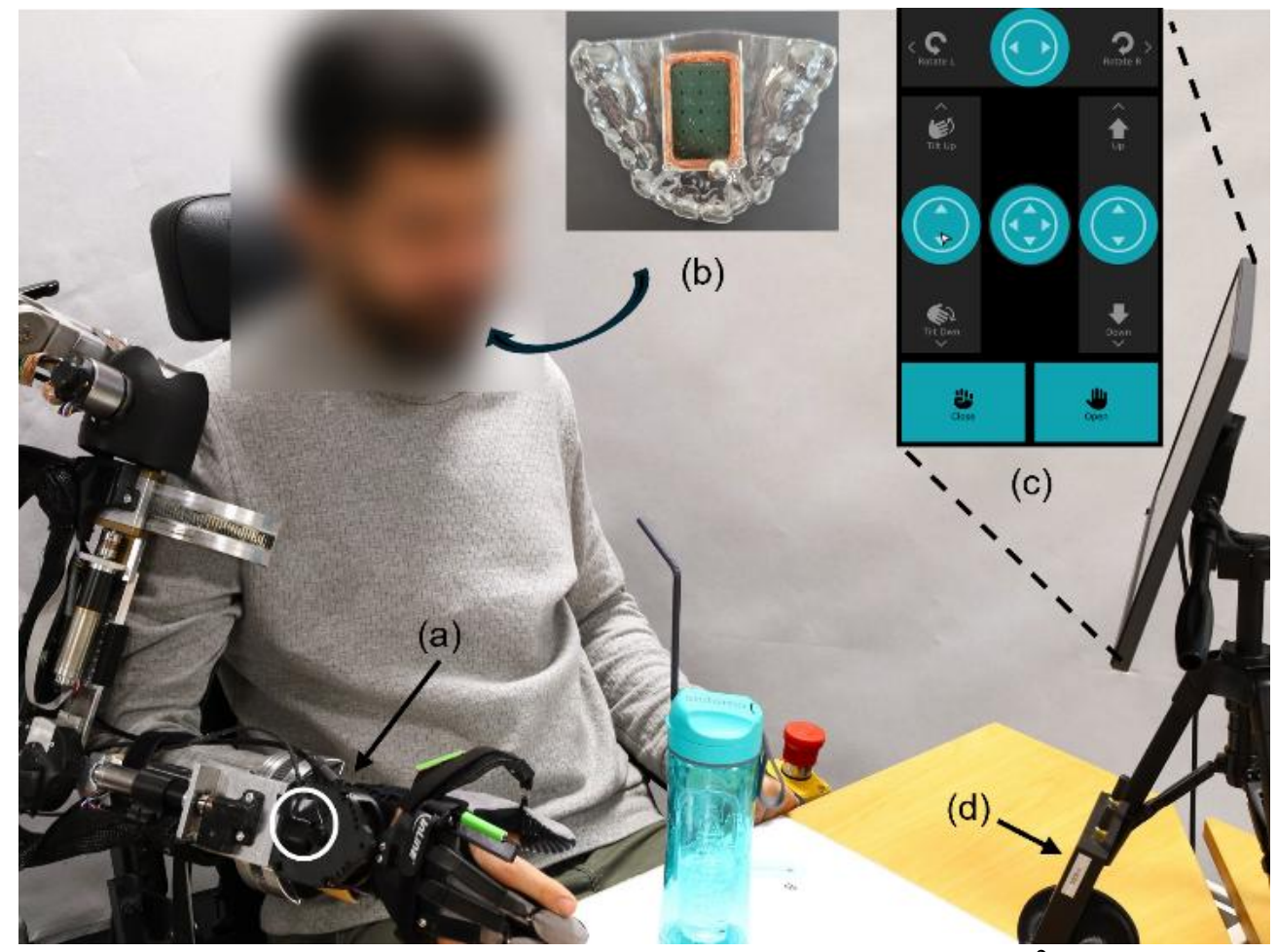

Fig 1. System Overview of the tongue-controlled EXOTIC$^{2}$ exoskeleton during experimental evaluation: (a) Wrist Ab-Ad joint in EXOTIC$^{2}$ with encoder position marked by white circle (b) nTCI mouthpiece unit glued to dental sheet (c) visual feedback of the tongue control layout displayed to participant on a screen and (d) nTCI central unit

The arm is supported by passive braces at the upper arm and forearm, while only the wrist and hand are directly coupled to the device[18]. This approach simplifies donning/doffing, accommodates user-specific anatomy, and reduces the risk of joint misalignment during use [18]. EXOTIC$^{2}$ weighs approximately 3.9 kg (including lightweight tendon-driven wrist Ab-Ad joint of about 0.2 kg), with wrist actuation components positioned proximally in a motor box to minimize distal loading and bulkiness of EXOTIC$^{2}$. The maximum hand linear velocity was set to 4 cm/s and wrist angular velocity to 0.2 rad/s for user safety and smooth motion based on pilot tests. Additionally, the mechanical limits of the wrist Ab-Ad joint can be adjusted based on each participant's pre-measured joint range of motion (ROM) limits. This design feature adds additional safety to avoid overstretching the joint. Furthermore, there were two buttons to cut power to the exoskeleton in case of emergencies as an additional layer of safety during experimental evaluation.

To assess the contribution of the Ab-Ad joint to functional task performance during ADLs, we evaluated the system with and without activation of the joint. With the activation of wrist Ab-Ad joint, an automatic leveling controller maintains the orientation of the hand frame in the global reference frame during hand motion, particularly useful in the drinking task for leveling the cup to avoid spill [15].

### 2) Non-Invasive Tongue-Robot Interface

The exoskeleton was controlled using a non-invasive tongue-robot interface (nTRI) [22]. The interface consisted of an intraoral mouthpiece unit and a central unit as shown in Fig. 2. The mouthpiece unit is mounted by means of a custom-fabricated removable dental appliance on the upper palate. Each dental appliance is individually produced using thermoforming a dental sheet (Duran, Scheu-dental GmbH, Germany) over a 3D-printed model of the participant's palate to ensure fit and stability.

The mouthpiece incorporated a sensor pad with 15 inductive sensors, integrated with electronic components for signal processing and wireless communication[22]. It also includes a rechargeable battery and an inductive-charging coil, and the assembly was enclosed in dental acrylic for safe and durable use. An integrated stainless-steel wireframe supported a titanium sphere, manipulated by user's tongue to either hover around the frame or made to touch the sensor pad (activate a command)[22]. The position of the point of activation on the sensor plate was obtained using the Mean Average of Neighbor Sensors method on the output of the 15 sensor coils [24]. The normalized position of the activation was then mapped to widgets on a graphical user interface displayed in front of the user for real-time visual feedback, enabling interaction like a stylus-based interface [24] (Fig. 1(c)).

Users controlled the system by positioning the metallic sphere over specific regions of the sensor pad, activating the widgets within the layout, which then generated commands for the exoskeleton. The layout consisted of three one-dimensional sliders and a two-dimensional joystick for proportional velocity control of six DoFs in EXOTIC[2] (Fig 1 (c) and 2 (b) and (c)), along with two virtual buttons for hand opening and closing. Selected widgets expanded after activation to support precise interaction (Fig. 2 (b) and (c)), and a dwell time of one and a half second was used to avoid unintended inputs.

## B. Experimental Setup and Procedures

*1) Participants:* Six individuals participated in this study, including one individual with ALS and five with SCI. Detailed participant information is presented in Table I. All participants had upper-limb impairments affecting their ability to independently perform ADLs. All participants had prior exposure to the tongue control interface through two earlier training sessions, ensuring familiarity with the control scheme before the evaluation. The study was approved by the regional ethics committee (The Scientific Ethics Committee for the North Jutland region, VEK, Protocol number: 20230044), and written informed consent was obtained from all participants prior to participation.

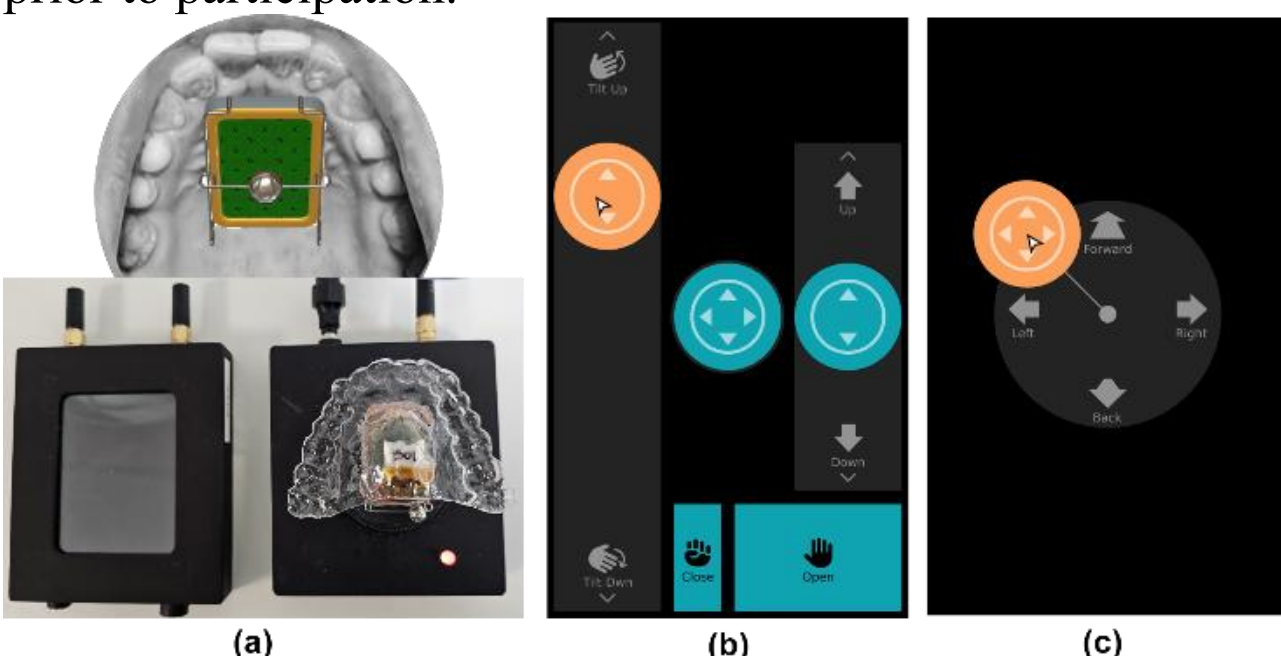

Fig. 2 (a) Non-invasive tongue computer interface: Top: in mouth demonstration of the mouthpiece unit with sensor pad and the wire frame. Bottom: Mouthpiece unit placed on charger and central unit, (b) Expansion of the vertical sliders after selection for fine control and (c) expansion of the central widget after selection and emulating a two-dimensional joystick.

*2) Experimental Procedure:* Each participant completed a single experimental session lasting approximately two hours. A within-subject experimental design was employed, in which each participant performed tasks under two conditions: (1) Wrist enabled condition, where the wrist Ab-Ad joint was active, allowing free motion with both manual control and automatic leveling; and (2) Wrist disabled condition, where the wrist joint was fixed in a neutral position by disabling the motor driver through software (Fig. 1(a) showing neutral wrist position). The order of conditions was randomized, and participants completed one familiarization trial same as actual experimental trial prior to data collection. Active or passive wrist Ab-Ad range of motion (ROM) was measured for each participant during a prior study on wrist ROM measurement using a digital goniometer[11]. Active ROM was measured only for participant P6 (SCI), whereas for rest of the participants (1 with ALS and 4 with SCI) passive ROM was assessed with assistance from the experimenter due to inability to actively perform wrist deviation due to motor impairment. These measurements were used to define individual joint limits in the exoskeleton, both applied from software and using mechanical limits in the joint.

### 3) Functional Tasks

Each participant performed two functional tasks, drinking and leveling representing common ADLs selected from prior user-centered studies and end-user desires collected using interviews and User-Board meetings at Center for rehabilitation Robotics, CREROB, at Aalborg University, with individuals with ALS and SCI [25], [26]. Each participant performed three trials of each functional task in both conditions (enabled and disabled wrist Ab-Ad joint). Trials that could not be finished due to object grasp failures were considered failed trials.

a) *Drinking task*: The object to grasp was a cup with water and a straw mounted within a secondary container with calibration markings(bottle) to prevent liquid spilling on the user or experimental setup (Fig 3). The calibration marks served

TABLE I STUDY PARTICIPANT INFORMATION

| Participant | Diagnosis | Sex | Age (years) | Height (cm) | Weight (kg) | ALSFRSr* | SCIM Score | AISA | Level of SCI | Time since diagnosis |
|---|---|---|---|---|---|---|---|---|---|---|
| P1 | ALS | M | 74 | - | - | 36 | N/A | N/A | N/A | 1yr 3m |
| P2 | SCI | M | 55 | 185 | 83 | N/A | - | AIS-B | C4 | 35yr |
| P3 | SCI | F | 53 | 169 | 82.8 | N/A | 17 | AIS-A | C3 | 1yr 6m |
| P4 | SCI | M | 22 | 167 | 39.7 | N/A | 24 | AIS-B | C5 | 9m |
| P5 | SCI | F | 30 | 179 | 63 | N/A | 23 | AIS-A | C4 | 6yr 3m |
| P6 | SCI | M | 56 | 183 | 87 | N/A | 90 | AIS-D | C6 | 19yr 6m |

(-): data missing (N/A): indicates not applicable (*):Abbreviation ALSFRSr - ALS Functional Rating Scale -Revised

to measure spillage. Participants started from the home position of the EXOTIC[2] (Fig. 3 (a)), reached for the cup at Position A on a table (Fig. 3 (c)). After grasping the cup, they moved it close to their face, touched the tip of the straw with their lips (Fig. 3 (b)) and placed it back at a designated location (Position C) on the table before releasing it. Spillage was assessed from post-trial images of the container (photos and videos of them used in each trial) and classified as *spill* if visible liquid residue or droplets were present, no-*spill* otherwise and failed trial when task couldn't be completed.

b) *Leveling scratch stick task*: Participants were provided with a cylindrical scratching stick positioned in the exoskeleton hand at the start of the trial at home position (Fig. 3 (a)). They were required to move the stick (Fig. 3 (b)) to a predefined location (Position C) on the table (Fig. 3 (c)) and place it in a leveled orientation before releasing it. Outcomes were categorized as: stick successfully placed on table defined as *leveled*, when stick fell defined as *not leveled* and when task couldn't be completed defined as failed trial.

## C. Outcome Measures

System performance was evaluated using a combination of kinematic and task-level metrics derived from recorded joint and control signals as detailed below:

### 1) Joint kinematics and task performance

Wrist abduction-adduction joint angles, user input signals, and task timestamps were recorded continuously at 100 Hz using the exoskeleton's onboard control system, with joint angles measured via integrated position sensors and user inputs logged from the tongue control interface. Task completion time (TCT) was computed for each trial, defined as the duration from task initiation (defined as first input given by user for movement initiation) to end (defined as last input given by user to release of objects). Task-specific performance metrics were recorded for each trial, which consisted of liquid spill during the drinking task and successful leveling of the scratching stick during placement.

### 2) Wrist ROM during task execution

For wrist-enabled trials, directional wrist usage was quantified from Ab-Ad trajectories. The following metrics were extracted per trial from encoder (Fig 1(a)) data: Abduction peak (°) defined as: maximum positive deviation, Adduction peak (°) defined as: maximum negative deviation, Task ROM (°) defined as: total ROM using sum of abduction and adduction joint angles. These values were summarized at the participant level using the maximum observed across trials. Additionally, baseline/ wrist ROM limits of each user were measured without robotic assistance as mentioned in experimental procedure.

### 3) Event-based wrist orientation

Wrist angles were extracted at key task events (drinking and leveling scratch stick) identified from user input signals, with grasp corresponding to hand closure and release to hand opening, detected by user input of grasp and release. For the drinking task, both grasp and release were analyzed, whereas for the scratch stick task only release was considered due to pre-positioning of the stick in user's hand

### 4) Usability Questionnaire

A custom questionnaire was used to capture participant-reported feedback on usability, comfort, and perceived functional benefit of the exoskeleton in an assistive context

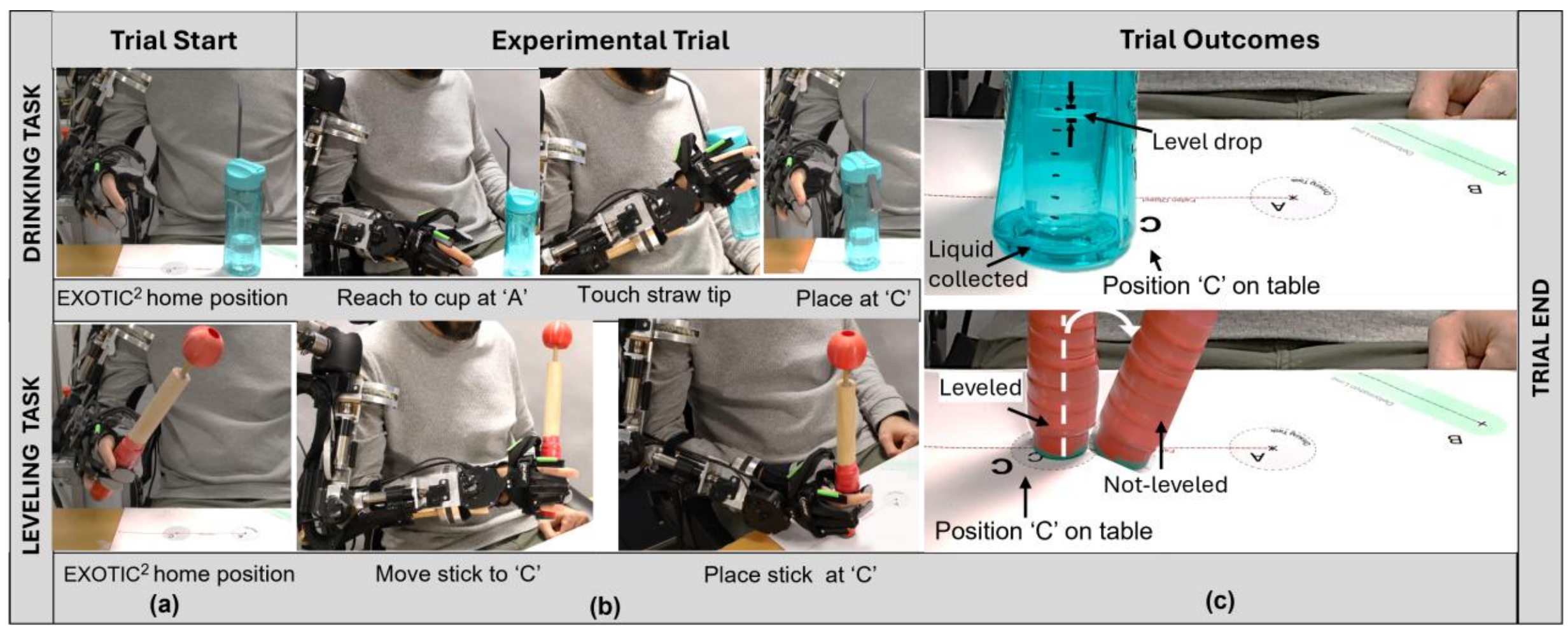

Fig 3. Experimental tasks demonstration (a) Trial start at home position, (b) experimental trial for each task and (c) Trial outcomes in each task, Top trial outcome: shows drinking task with the cup (inside bottle) placement position on table and Bottom trial outcome: shows Scratch stick task, with stick placement position on table

(Table II). The questionnaire comprised 14 questions (Q1-Q14) grouped into four themes: (i) comfort and physical interaction (Q1-Q4), (ii) perceived functional impact of wrist abduction-adduction (Q5-Q6), (iii) usability and ergonomics (Q7-Q12), and (iv) user acceptance (Q13-Q14) (Table II). All items were rated on a 10-point Likert scale, where 1 represented the most favorable outcome and 10 the least favorable (Table II). For Q1-Q3, participants provided ratings under both experimental conditions (with and without wrist abduction-adduction), enabling within-subject comparison. Q5-Q6 assessed perceived task-level improvement with the wrist joint, while Q4 and Q7-Q12 captured general usability aspects such as size, weight, donning/doffing, and comfort. Q13-Q14 evaluated willingness to use the system with and without the wrist joint.

The questionnaire was administered after the experimental session, and additional comments were collected to support interpretation. Descriptive statistics of responses were analyzed, including median and interquartile range, and interpreted alongside objective performance measures to provide complementary insight into user experience.

### D. Data Analysis

Experimental data was processed offline in Python. Signals were synchronized using timestamps at 100 Hz, and task events were detected from thresholds in the exoskeleton glove open/close events. Trial-level metrics were summarized at the participant level for descriptive analysis presented in the results section. Due to the small sample size and inclusion of a single ALS participant, results are reported without inferential statistics, focusing on participant-wise trends and condition-based comparisons.

## III. RESULTS

All participants completed the drinking and scratch stick tasks with and without the wrist Ab-Ad joint active. The detailed findings for task performance, kinematic analysis, and user feedback are presented below:

### A. Quantitative Performance Outcomes

#### 1) Drinking Task

Enabling wrist abduction/adduction substantially improved task outcomes (Fig4 a). The repetition of trials completed without spill increased from 16.7% (without wrist Ab-Ad) to 72.2% (with wrist Ab-Ad). The rate of failed trials remained low and unchanged across conditions (5.6%).

Task completion time decreased in 4 out of 6 participants in the drinking task when the wrist Ab-Ad DoF was enabled, with reductions ranging from approximately 10 to 75 s, while the remaining participants demonstrated comparable performance (Fig. 4d).

#### 2) Scratch stick leveling task

. The successful placements (leveled stick) during scratch stick leveling task increased from 16.7% (without wrist Ab-Ad) to 77.8% (with wrist Ab-Ad) (Fig. 4 b). Failed trials also reduced from 16.7% to 5.6% when wrist Ab-Ad joint assistance.

Task completion times in the scratching task were generally shorter than in the drinking task (because of task simplicity),

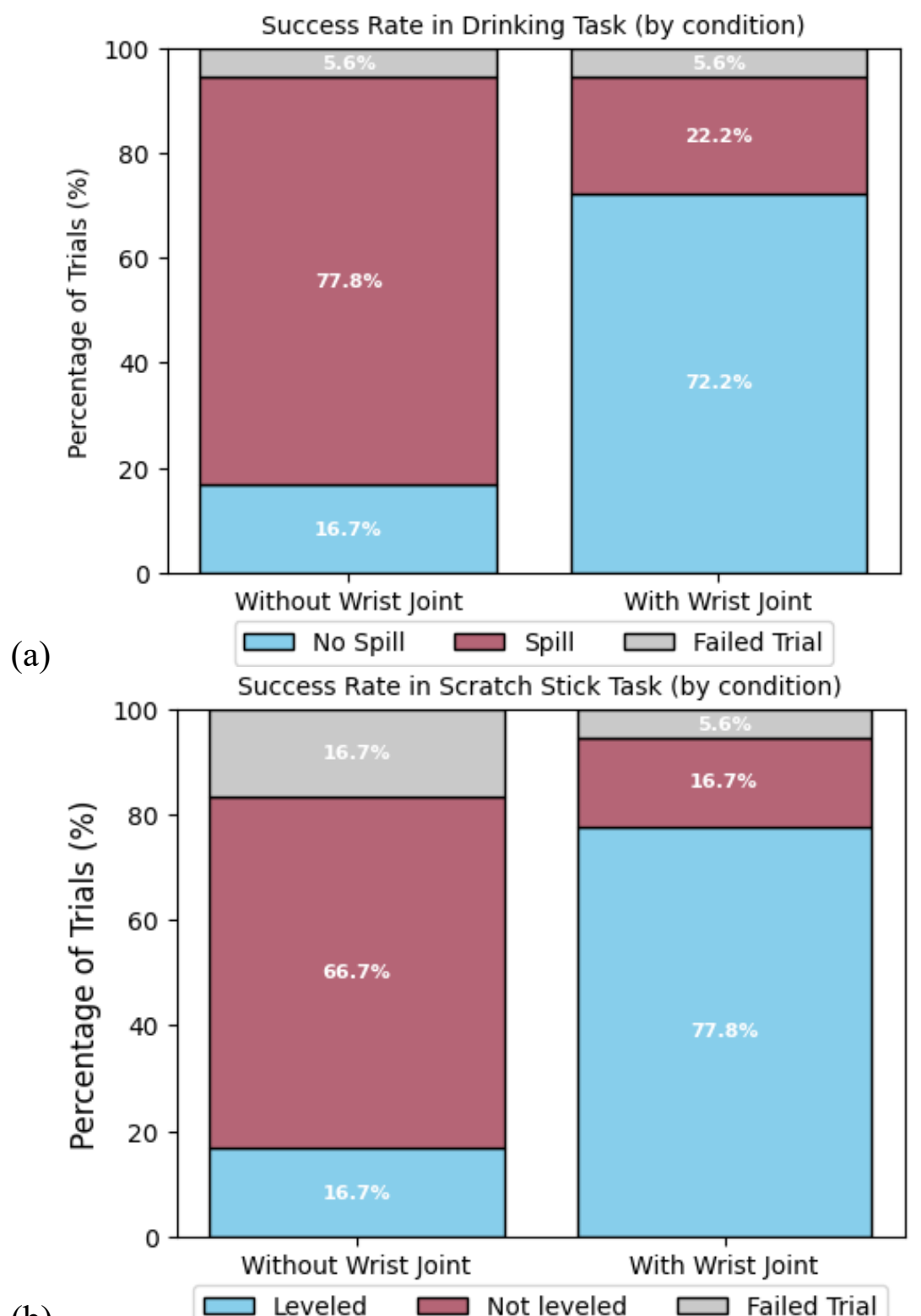

Fig. 4 Task outcomes during experimental trials (a) Spill outcomes during drinking task, (b) leveling outcomes during scratch stick task

showed small improvements with wrist Ab-Ad assistance across participants (approximately 0-25 s) and more variable across participants, with no consistent trend observed (Fig. 5).

### B. Wrist Usage and Kinematics

#### 1) Event-Based Wrist Orientation

In the drinking task, wrist angles at grasp and release revealed consistent directional usage (Ab-Ad trajectory) when the wrist DoF was enabled (Fig. 6 a). Across participants with SCI, grasp angles were typically in the adduction range, while release angles shifted toward abduction, indicating active reorientation of the end-effector during object manipulation (Fig,6. a). The participant with ALS exhibited a similar directional trend but with reduced variability and a narrower range of wrist usage (Fig. 6 a). In the scratching task, release angles were clustered closer to a consistent orientation across participants, with most participants with SCI achieving positive (abducted) wrist angles at release (Fig 6. b). This reflects a strategy to align the stick with the table surface during placement. The participant with ALS demonstrated higher release angles (approximately 18°) with greater dispersion (Fig. 6 b).

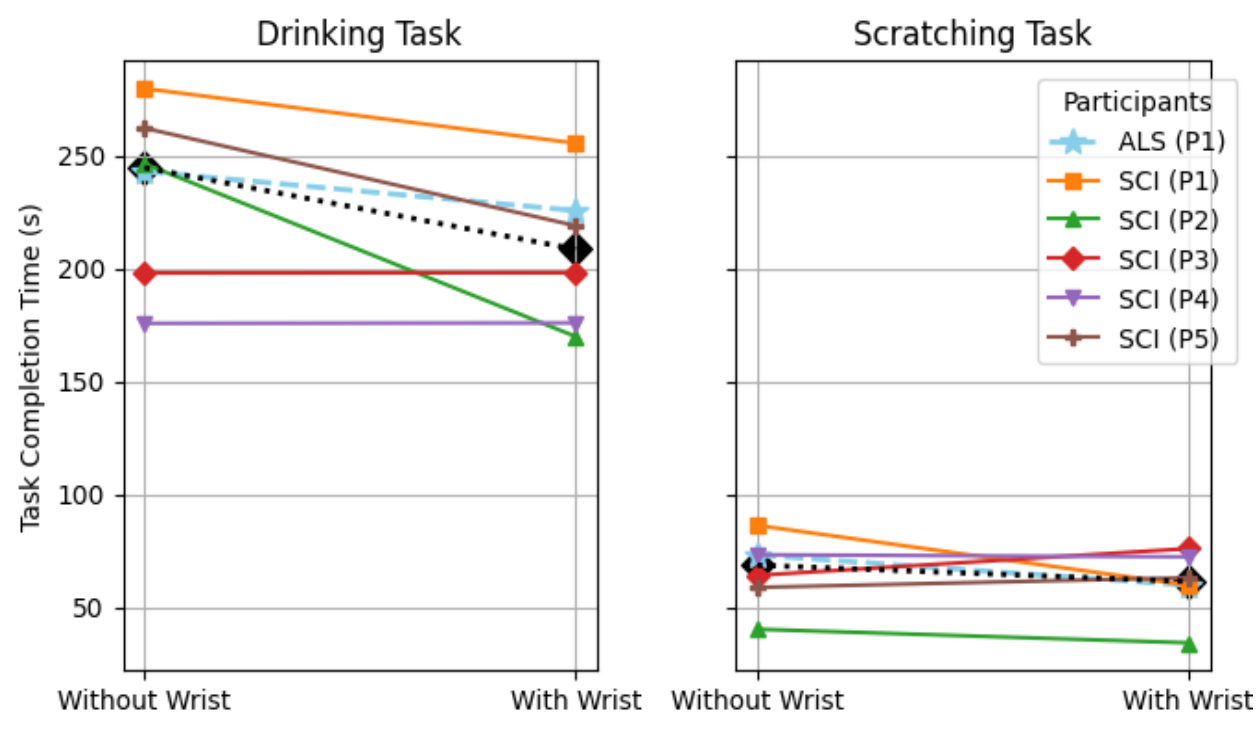

Fig. 5 Task Completion time (TCT) during drinking and scratch stick task for both ALS and SCI participants.

### 2) *Wrist Ab-Ad ROM Utilization*

A comparison of participant's wrist ROM limits with task-executed ROM showed that participants utilized a subset of their available motion during task performance (Fig 6. c). Across participants, experimental abduction peaks were consistently lower than their ROM limits, while adduction usage was similarly reduced in magnitude (Fig 6. c). This indicates that task demands did not require full exploitation of the available joint range. Task-dependent differences were observed, with drinking generally involving larger abduction excursions than scratching. Variability in ROM usage across participants suggests individualized control strategies.

## *B. Qualitative outcomes and User Feedback*

Participant-reported feedback complemented the quantitative findings by providing insight into physical interaction, perceived task benefit, and overall usability of the system. Questionnaire responses were analyzed across four themes: physical comfort, task-level functional impact, system usability, and user acceptance. Across all participants, physical comfort ratings remained consistently favorable. With rating from 1 to 4 in a scale of 1 to 10, with 1 being most favorable, Scores for interface-related and task-related discomfort (Q1-Q3) were in the range 1-4 and 1-2, respectively and were low and unchanged between conditions (wrist Ab-Ad active vs locked), indicating that enabling wrist Ab-Ad did not introduce additional discomfort during use. Similarly, extended-use comfort (Q4) was rated positively (Table II).

Perceived functional impact (Q5-Q6) reflected task-dependent benefits consistent with the objective results (Fig 7. a). Participants reported clear improvement in the drinking task, with low scores indicating strong perceived benefit. The responses for the scratching task showed greater variability, suggesting that the added degree of freedom was beneficial but not uniformly required across all users (Fig. 7 a). Additionally, all participants reported that the addition of the wrist Ab-Ad joint did not worsen performance in either the drinking or scratching tasks, with all responses at the lowest score (1).

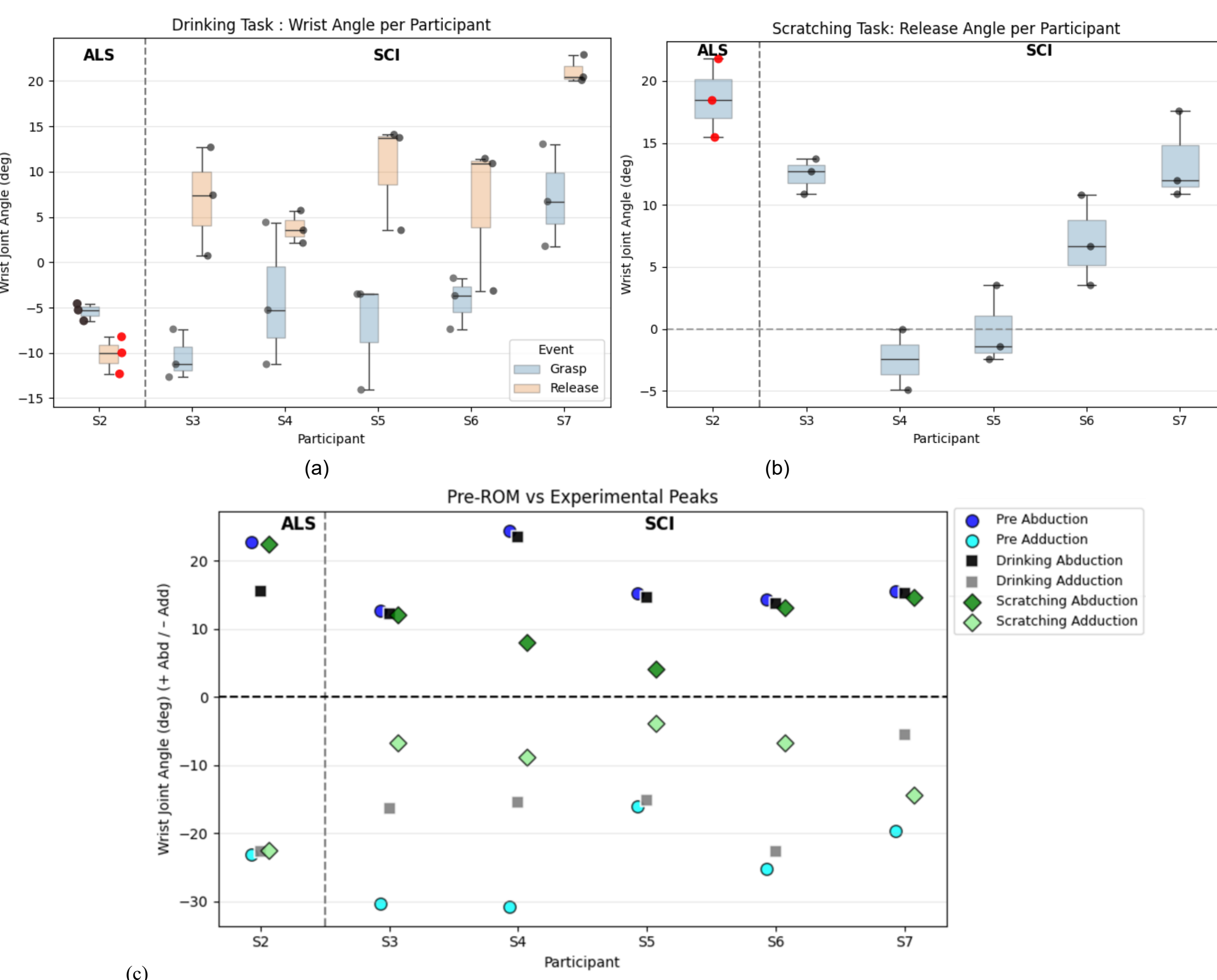

Fig. 6 Wrist Ab-Ad usage and kinematic outcomes (a) event-based wrist joint-angle distribution during drinking task, (b) event-based wrist joint-angle distribution during scratch stick task, and (c) Participant's ROM and experimental ROM during both drinking and scratch stick task

TABLE II
MEDIAN AND RANGE OF PARTICIPANT RESPONSES TO QUESTIONNAIRE ITEMS Q1-4 AND Q7-8 (1: MOST FAVORABLE AND 10: LEAST FAVORABLE)

| Theme | Description | Median | Range | Interpretation |
|---|---|---|---|---|
| Comfort and Physical Interaction | Q1:Tongue interface discomfort (with/without wrist) | 1 | 1-4 | No additional discomfort observed due to wrist Ad.- |
| | Q2: Drinking task discomfort (with/without wrist) | 1 | 1-2 | No discomfort during task |
| | Q3: Scratching task discomfort (with/without wrist) | 1 | 1-2 | No discomfort during task |
| | Q4: Pressure/ discomfort | 2.75 | 1-4 | Generally comfortable, minor issues |
| Usability & Ergonomics | Q7: Perceived size | 2.5 | 1-6 | Acceptable size, some variability |
| | Q8: Perceived weight | 1 | 1-2 | Consistently lightweight |
| | Q9: Donning/ doffing ease | 3.5 | 1-8 | Variable, usability challenge identified |
| | Q10: Ergonomic design | 4 | 1-10 | Mixed perception across users |

Usability-related responses (Q7-Q10) indicated generally favorable system characteristics (medians scores in range 1-4), particularly in terms of weight and size, which were consistently rated favorably (median 1 and 2.5 respectively (Table II)). However, responses related to donning and doffing (Q9) and overall ergonomic design (Q10) showed greater variability across participants, with most reporting moderate scores while two out of six users rated these aspects highly (score = 1)(Fig 7 c).

Finally, responses related to perceived value and acceptance (Q11, Q12, Q13-Q14) indicated that participants recognized the functional usefulness of the wrist joint. The added wrist Ad-Ab joint increased system acceptance in three of the six participants, decreased it in one and remaining three were unchanged (Fig. 7 d). Two participants (P1 and P5), who rated with wrist acceptance lowest, mentioned that their acceptance of the system was lower because of the control interface. which resulted in a rating of 9 for P1 and P5 while the others had acceptance rates between 1 and 5. Scorings for using the system without the wrist joint were lower overall, suggesting that the additional degree of freedom was beneficial in most cases.

## IV. DISCUSSION

This study presents a user-centered evaluation of wrist Ab-Ad in an assistive upper-limb exoskeleton, focusing on its functional impact during representative ADLs. By combining task-level performance metrics with user feedback, the results provide insight into the clinical relevance of the wrist Ab-Ad DoF beyond purely kinematic capability. The results demonstrate that wrist Ab-Ad contributes directly to improved ADL execution. Enabling the wrist Ab-Ad DoF led to clear increase in task success across both drinking and scratching tasks, indicating improved end-effector alignment and control during object interaction. Across both tasks, enabling wrist Ab-Ad resulted in: improved task success rates, reduced reliance on compensatory strategies (e.g., excessive repositioning) and more consistent wrist orientation at task-critical events. While inter-participant variability was present, the overall trend was consistent across participants with SCI. The participant with ALS demonstrated qualitatively similar behavior, with differences in variability and execution patterns.

These findings are consistent with prior work emphasizing the role of distal joints in a ULE such as wrist Ab-Ad in functional manipulation, where wrist orientation enables more precise interaction with the environment [27], [28]. Importantly, the objective improvements observed here are supported by participant-reported outcomes, with users perceiving better task performance when wrist Ab-Ad joint was active, particularly in the drinking task. The agreement between quantitative metrics and subjective feedback suggests that the added wrist Ab-Ad DoF provided meaningful functional benefit rather than a purely kinematic enhancement.

Many participants used only a portion of their available wrist ROM, especially wrist adduction during task execution. This indicates that functional performance depends less on maximizing joint range and more on the ability to access and control task-relevant regions of motion. This observation aligns with prior work in rehabilitation robotics, where effective control and usability often outweigh raw mechanical capability [29], [30]. From a design perspective, this suggests that increasing ROM beyond task requirements may provide limited benefit unless it remains easily controlled.

The improvements in task performance were achieved without increasing perceived task complexity, suggesting that the additional wrist Ab-Ad DoF was intuitive within the applied control interface. Participants did not report increased discomfort with the wrist enabled, indicating that the added wrist Ab-Ad functionality did not negatively affect physical interaction or system tolerability. At the same time, variability in user feedback, particularly regarding donning/doffing and ergonomics highlights practical challenges not captured by performance metrics alone. This suggests that, although usability was favorable for some users, further refinement may improve consistency in system setup and fit for real-world adoption of ULEs [4], [31]. Importantly, none of the participants perceived a deterioration in task performance with the addition of the wrist Ab-Ad joint, suggesting that the added DoF does not introduce functional drawbacks.

Inter-participant differences in wrist usage and user acceptance further emphasize the need for adaptable system design. While all the participants recognized the functional advantage of the wrist joint, willingness to use the system varied, suggesting that perceived benefit depends not only on performance but also on usability, effort, and individual

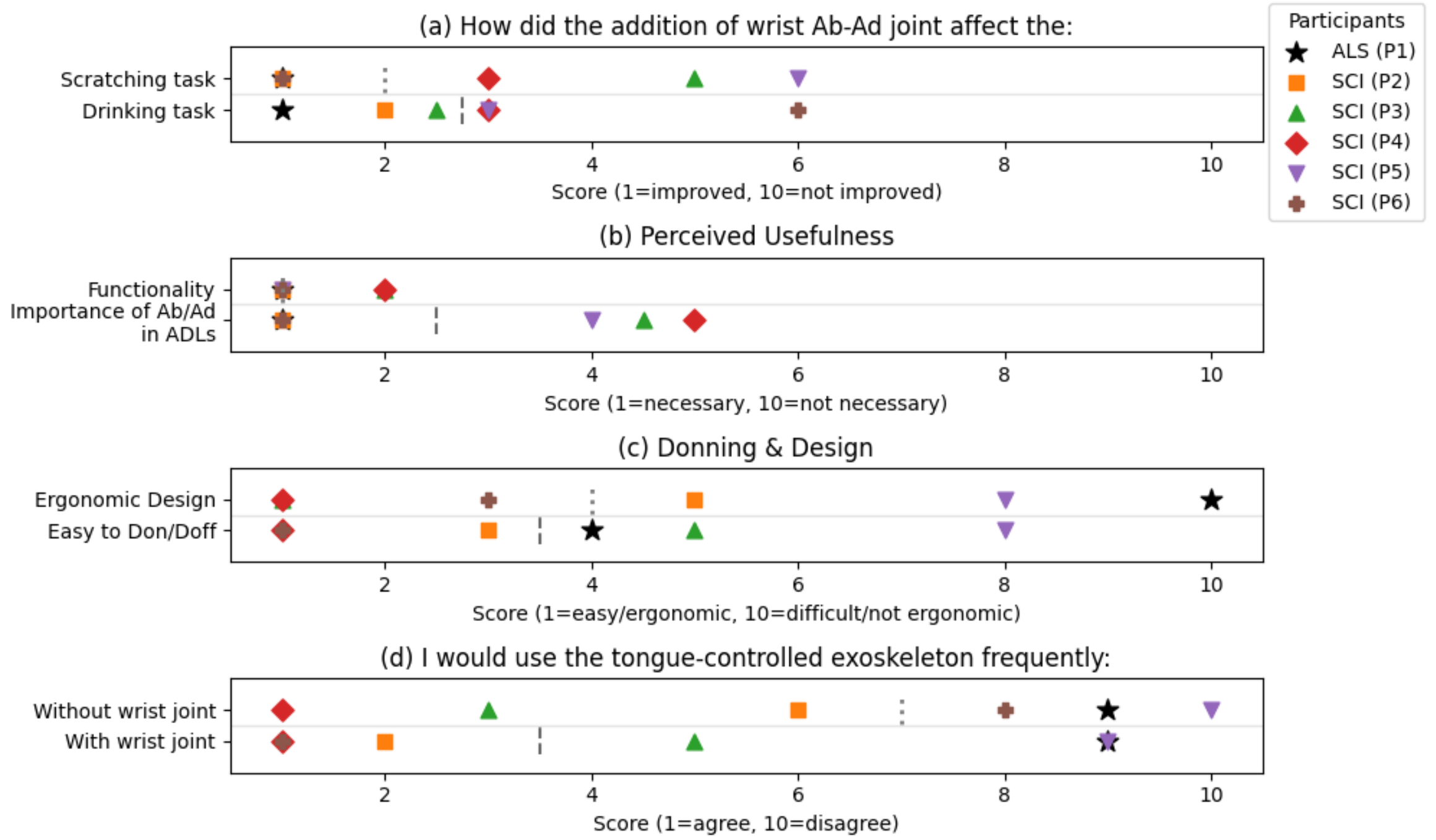

Fig. 7 Qualitative feedback from system usability questionnaire of each participant (a) Responses from questions 5 and 6, (b) Responses from questions 11 and 12, (c) Responses from questions 9 and 10 and (d) Responses from questions 13 and 14

preference [32]. The inclusion of a participant with ALS provides an initial indication of applicability across different impairment profiles. Although limited to a single case, the observed differences in execution patterns suggest that impairment-specific factors may influence control strategies and should be considered in future work.

This is an exploratory study, limited by a small sample size and the inclusion of a single ALS participant restricts generalizability. As such, the analysis remains descriptive, and larger studies are required to confirm these findings. Further, some of the questions in the questionnaire did not discriminate between the control interface, the whole exoskeleton, and the wrist Ab-Ad joint which means some answers may also relate to other aspects as the wrist Ab-Ad joint. The results indicate that wrist Ab-Ad can substantially enhance functional task performance, but its effectiveness depends on both task context and usability. While the added wrist Ab-Ad DoF improved objective performance, variability in user feedback highlights the importance of balancing functionality with practical usability.

Future designs should therefore prioritize task-relevant and controllable ROM, intuitive integration of additional DoFs, support of task-specific interaction strategies, setup complexity and donning/doffing. These considerations are critical for translating technical capability into clinically meaningful assistive function. Future work should also include long-term evaluations, as short study duration may limit mastery of the tongue interface and affect user acceptance; extended use has shown up to 30% performance improvement [33]. The nTCI, closely related to the TCI [34], [35], has been used to control computers and powered wheelchairs [23], [35], [36] and is commercially available, enabling multi-device control through a single interface.

## VI. Conclusion

This exploratory study provides a user-centered clinical evaluation of wrist abduction-adduction in an assistive upper-limb exoskeleton during representative ADLs performed by users with tetraplegia. The results show that enabling the wrist Ab-Ad degree of freedom improves task success, particularly in tasks requiring precise alignment, without increasing user discomfort. At the same time, variability in usability and acceptance highlights that functional improvements must be considered alongside practical factors such as ergonomics, setup complexity, and individual user strategies.

The work offers clinical evidence that wrist Ab-Ad can enhance assistive function in individuals with severe motor impairments and the results reinforce that clinically meaningful assistive performance depends on the integration of functional capability with user-centered design.

## Acknowledgment

This work has been funded by the Aage and Johanne Louis-Hansens Foundation, (Grant No. 20-2B- 7273), Denmark at Aalborg University. We thank Irina Simona Socol-Lontis for assistance during experiments.